\documentclass[runningheads]{llncs}
\usepackage{float}
\usepackage{graphicx}
\usepackage{float}
\usepackage{subfigure}
\usepackage{float}
\restylefloat{table}
\usepackage{placeins}
\usepackage{amsmath}
\floatstyle{plaintop}
\restylefloat{table}
\usepackage{multirow}
\usepackage{graphicx}
\usepackage{amssymb}
\usepackage{color}
\usepackage[hidelinks]{hyperref}
\usepackage{hyperref}
\usepackage{marvosym}

\usepackage[doipre={doi:~}]{uri}
\begin{document}
\title{Detect Any Deepfakes: Segment Anything Meets Face Forgery Detection and Localization}


\titlerunning{Detect Any Deepfakes}
%
%

%
\author{Yingxin Lai \inst{1}, Zhiming Luo \inst{1} \and
Zitong Yu\textsuperscript{2\thanks{Corresponding author}}}

\institute{Xiamen University \and Great Bay University }

\maketitle              
\begin{abstract}
The rapid advancements in computer vision have stimulated remarkable progress in face forgery techniques, capturing the dedicated attention of researchers committed to detecting forgeries and precisely localizing manipulated areas. Nonetheless, with limited fine-grained pixel-wise supervision labels, deepfake detection models perform unsatisfactorily on precise forgery detection and localization. To address this challenge, we introduce the well-trained vision segmentation foundation model, i.e., Segment Anything Model (SAM) in face forgery detection and localization. Based on SAM, we propose the Detect Any Deepfakes (DADF) framework with the Multiscale Adapter, which can capture short- and long-range forgery contexts for efficient fine-tuning. Moreover, to better identify forged traces and augment the model's sensitivity towards forgery regions, Reconstruction Guided Attention (RGA) module is proposed. The proposed framework seamlessly integrates end-to-end forgery localization and detection optimization. Extensive experiments on three benchmark datasets demonstrate the superiority of our approach for both forgery detection and localization. The codes will be released soon at \href{https://github.com/laiyingxin2/DADF}{\textcolor{magenta}{https://github.com/laiyingxin2/DADF}}.

\keywords{Deepfake \and SAM  \and Adapter \and Reconstruction learning.}
\end{abstract}
\section{Introduction}
\vspace{-1.0em}


Amongst the diverse human biometric traits, the face is endowed with relatively abundant information and holds significant prominence in identity authentication and recognition. Nonetheless, with the rapid progress of computer vision technology, an array of face-changing techniques has emerged. In particular, the widespread usage of software such as FaceApp and FakeApp has drawn considerable attention to the field of face forgery detection \cite{li2020face}. Therefore, both industry and academia are in urgent need of robust detection methods to mitigate the potential misuse of face forgery technology. 
 
 \begin{figure}[t]
	\centering  
	\includegraphics[width=0.75\linewidth]{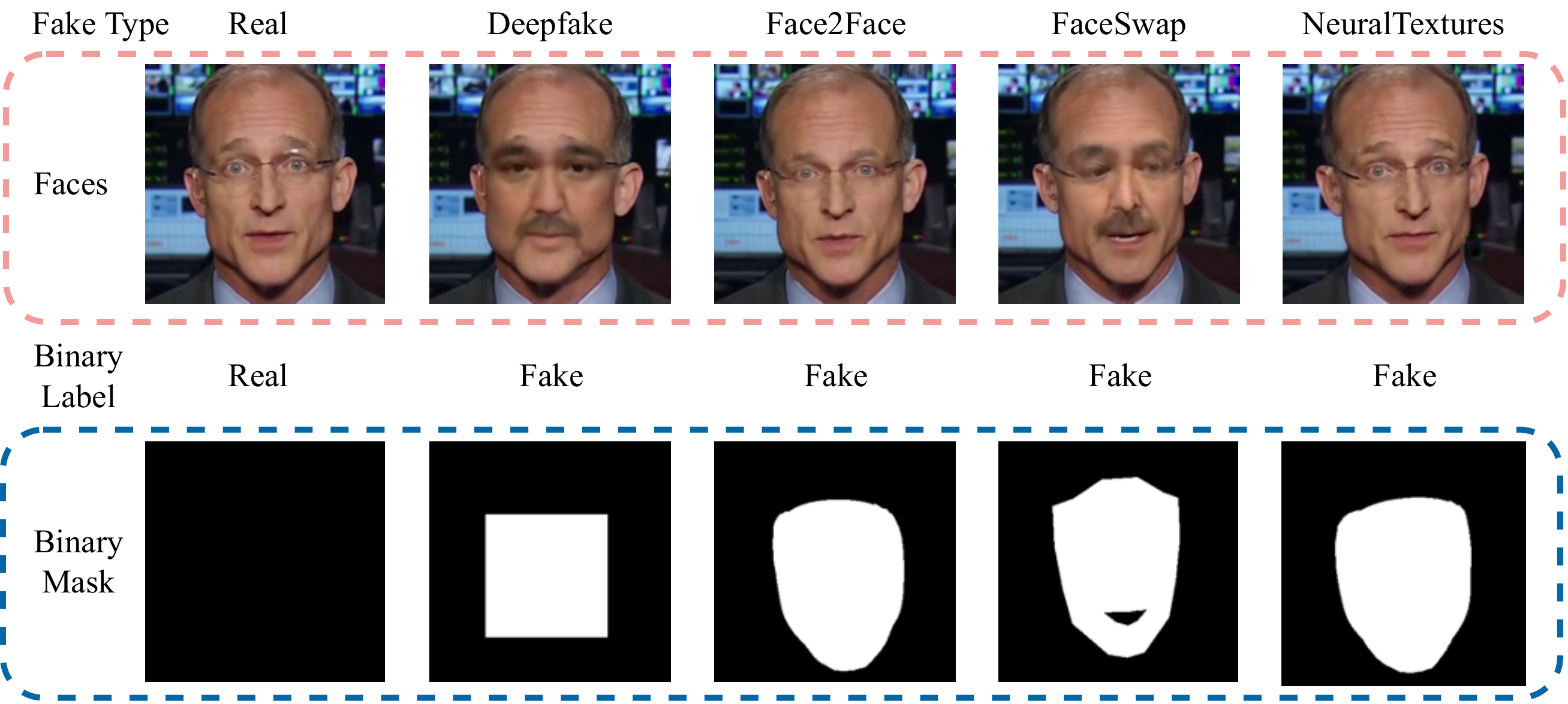}
    \vspace{-1.0em}
	\caption{Visualization of the real face and its four common forgery manners in FaceForensics++ \cite{rossler2018faceforensics}, as well as their binary labels and corresponding forgery areas.}
 \vspace{-1.3em}
 \label{fig1}
\end{figure}


Currently, the majority of forgery detection methods treat the task as a binary classification problem \cite{1,swaminathan2008digital,chierchia2011prnu,tan2019efficientnet} and utilize convolutional neural networks (CNNs) for feature extraction and classification. Although continuous advancements in forged face detection technology in recent years, accurate localization of forged regions remains a challenge, particularly for models that solely provide classification results. The precise identification of forged regions holds the utmost importance for uncovering the intentions and interpretability of the perpetrators. It allows individuals to discern fake images based on the forged regions and observe the discrepancies between forged and genuine images. Fig. \ref{fig1} illustrates four common forgery manners in FaceForensics++ \cite{rossler2018faceforensics} and their corresponding pixel-level masks, which are exhausted to be annotated. Due to the limited fine-grained pixel-wise forgery labels, some forgery localization methods \cite{kong2022detect,huang2022fakelocator,guo2023hierarchical,guan2023collaborative} trained from scratch usually suffer overfitting.\par

Recently, Meta introduces the pioneering foundational segmentation model, i.e., Segment Anything Model (SAM) \cite{kirillov2023segment,zhang2023comprehensive}, demonstrating robust zero-shot segmentation capabilities. Subsequently, researchers have explored diverse approaches such as Low-Rank Adaptation (LoRA) \cite{zhou2014low}, SAM adapter \cite{chen2023sam}, and learnable Prompt \cite{qiu2023learnable} to fine-tune SAM on downstream tasks including medical image segmentation and anomaly detection. However, these methods usually yield unsatisfactory positioning outcomes in face forgery localization due to their weak forgery local and global context modeling capacities.\par


This study focuses on 1) investigating how SAM and its variants perform in the deepfake detection and localization task; and 2) designing accurate and robust pixel-level forgery localization methods across various datasets. For the former one, we find that due to the limited multiscale and subtle forgery context representation capacity, SAM \cite{kirillov2023segment} without or with fine-tuning \cite{zhou2014low,qiu2023learnable} cannot achieve satisfactory forgery detection and localization results, which can be alleviated via the proposed SAM based Multiscale Adapter. On the other hand, we find that SAM and its variants are sensitive by the forgery boundary and domain shifts, which might be mitigated by the proposed Reconstruction Guided Attention module. Our main contributions are summarized as follows:
\begin{itemize}
\item[-] We are the first to explore the availability of SAM and its fine-tuning strategies in the deepfake detection area. Based on SAM, a novel and efficient Detect Any Deepfakes (DADF) framework is proposed. 
\item[-] We propose the Multiscale Adapter in SAM, which can capture short- and long-range forgery contexts for efficient fine-tuning.
\item[-] We propose the Reconstruction Guided Attention (RGA) module to enhance forged traces and augment the model's sensitivity towards forgery regions.
\item[-] The proposed method achieves state-of-the-art performance in terms of both forgery detection and localization.
\end{itemize}

\vspace{-1.3em}
\section{Related work}
\vspace{-0.5em}
\subsection{Face Forgery Detection}
\vspace{-0.5em}
\noindent\textbf{Binary classification.}
Face forgery detection is predominantly treated as a binary (real/fake) classification task. Plenty of deep learning based methods are developed for detecting face forgery. Li et al. \cite{liy2018exposingaicreated} observed distinct variations in blink frequency between forged and authentic videos. To exploit this disparity, the authors utilized CNNs and Long short-term memory (LSTM) \cite{hochreiter1997long} models to extract blink-related features in the time domain, enabling the classification of video authenticity. Agarwal et al. \cite{agarwal2021detecting} observed a mismatch between the movements of the mouth, ears, and chin in fake faces despite synchronized audio. Dang et al \cite{dang2020detection} incorporated an attention mechanism to emphasize the forgery area, leading to improved accuracy in forgery classification. Alternatively, Nguyen et al. \cite{nguyen2019capsule} proposed a capsule network designed specifically for identifying counterfeit images or videos. Additionally, Fang et al. \cite{nguyen2019multi} introduced the integration of reconstruction losses alongside classification to enhance the overall performance of the model. However, the above-mentioned methods only provide the result of forgery on the scale of the whole image, thus ignoring the identification of the forged region, which lack sufficient interpretability.

\vspace{0.3em}
\noindent\textbf{Joint detection and localization.}
Face forgery localization precisely identifies manipulated regions of a face at the pixel level. A hybrid CNN-LSTM model \cite{bappy2017exploiting} was proposed for learning the distinctive boundary variations between manipulated and non-manipulated regions.
Nguyen et al. \cite{nguyen2019multi} proposed to utilize multi-task learning to detect and locate manipulated regions in both images and videos. Attention mechanisms \cite{dang2020detection,kong2022detect} are used to enhance the feature maps of the classification task by highlighting the forged regions.
Li et al. \cite{li2020face} proposed to detect the edge regions with mixed boundaries between the manipulated face and the background. However, there are still no works investigating vision segmentation foundation models for joint face forgery detection and localization.

\vspace{-1.3em}
\subsection{Foundation Model}
\vspace{-0.5em}
In recent years, the development of large-scale deep learning pre-training models has promoted the rapid progress of basic visual models in various tasks in the computer field. Models such as BERT \cite{devlin2018bert} and GPT have improved abilities of language understanding, inference and generation in the field of natural language processing, and these models only need the Prompt of specific tasks to apply to new language tasks.
CLIP \cite{radford2021learning} uses contrast loss to learn large-scale image-text pairs. It achieves excellent classification performance in specific downstream tasks without additional data training.
COSTA \cite{mensink2014costa} combines the prior knowledge of various pre-training paradigms and unifies them to achieve the most advanced zero-shot classification.
DINOv2 \cite{oquab2023dinov2} can detect objects in the open world through a given text.
SAM \cite{kirillov2023segment} proposes a basic model based on image segmentation, using a diverse, high-resolution, licensed and privacy-protected 11 million images and 1.1 billion high-quality segmentation masks for training. The model can accept multiple prompts as input, such as points, boxes and prompts, and can extract high-quality target masks in open-world scenarios. Inspired by these basic models, we try to fine-tune the existing models to achieve the goal of face forgery detection and anomaly localization.

\begin{figure}[t]
\vspace{-1.3em}
\centering
\includegraphics[width=110mm]{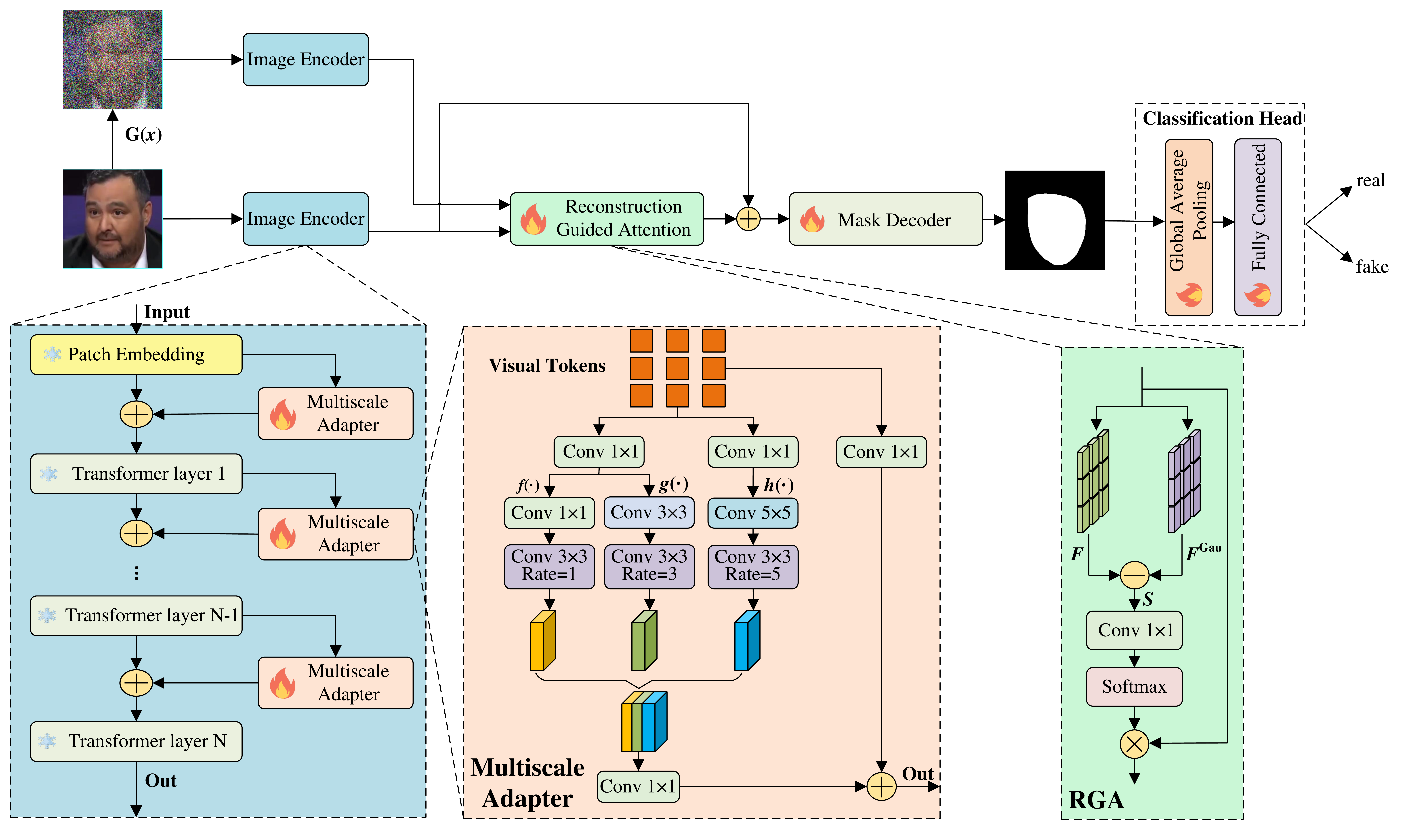}
\vspace{-1.3em}
\caption{\small{Framework of the proposed Detect Any Deepfakes (DADF). The Multiscale Adapters, Reconstruction Guided Attention module, mask decoder of SAM, and the classification head are trainable while the pre-trained SAM encoder is fixed.}}
\label{Fig.main1}
\vspace{-1.3em}
\end{figure}

\vspace{-1.3em}
\subsection{Parameter-Efficient Fine-Tuning}
\vspace{-0.5em}
The training cost of adjusting all parameters for foundation models when adapting to downstream tasks may be very high, along with great pressure on storage resources. To alleviate this problem, some researchers have developed efficient fine-tuning strategies. Liu et al. \cite{liu2021p} 
proposed continuous learnable prompts for the frozen language model, which significantly reduces the storage and memory usage required for downstream task adaptation.
Clark et al. \cite{clark2020electra} 
proposed to train a generative network to sample and replace the original input token, thus reducing the demand for computing resources while ensuring training accuracy.
Zhou et al. \cite{zhou2014low} proposed the Low Rank Adaptation (LoRA) to indirectly train few dense layers in transformer layers by optimizing the rank decomposition matrix of the dense layer changes during the adaptation process, while maintaining the weight of the pre-training unchanged. The most similar works to ours are the adapter-based fune-tuning \cite{chen2023sam,wu2023medical}. Instead of exploiting only channel-wise new knowledge for SAM, the proposed multiscale adapter is able to mine short- and long-range forgery contexts for efficient SAM fine-tuning.

\vspace{-1.3em}
\section{Methodology}
\vspace{-0.5em}

As illustrated in Fig. \ref{Fig.main1}, the proposed SAM-based \cite{kirillov2023segment} architecture involves an image encoder with the Multiscale Adapters for feature extraction, a Reconstruction Guided Attention (RGA) module for forged feature refinement and a mask decoder for forgery mask prediction. Based on the predicted mask, a classification head consisting of global average pooling and fully connected layers is cascaded for real/fake classification. 

\vspace{-1.3em}
\subsection{Multiscale Adapter for SAM}
\vspace{-0.5em}

\begin{figure}[t]
\vspace{-1.3em}
    \centering
    \subfigure[]{\includegraphics[width=0.23\textwidth]{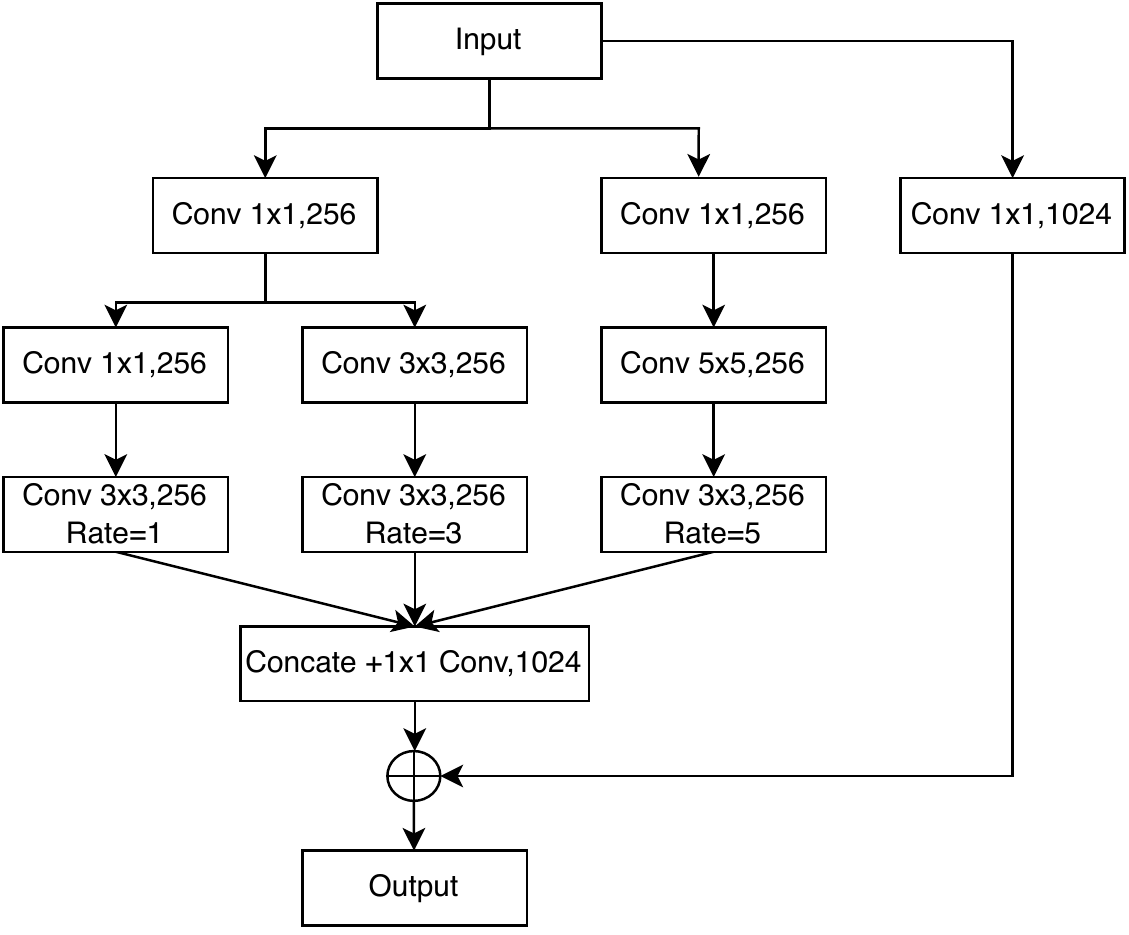}} 
    \subfigure[]{\includegraphics[width=0.26\textwidth]{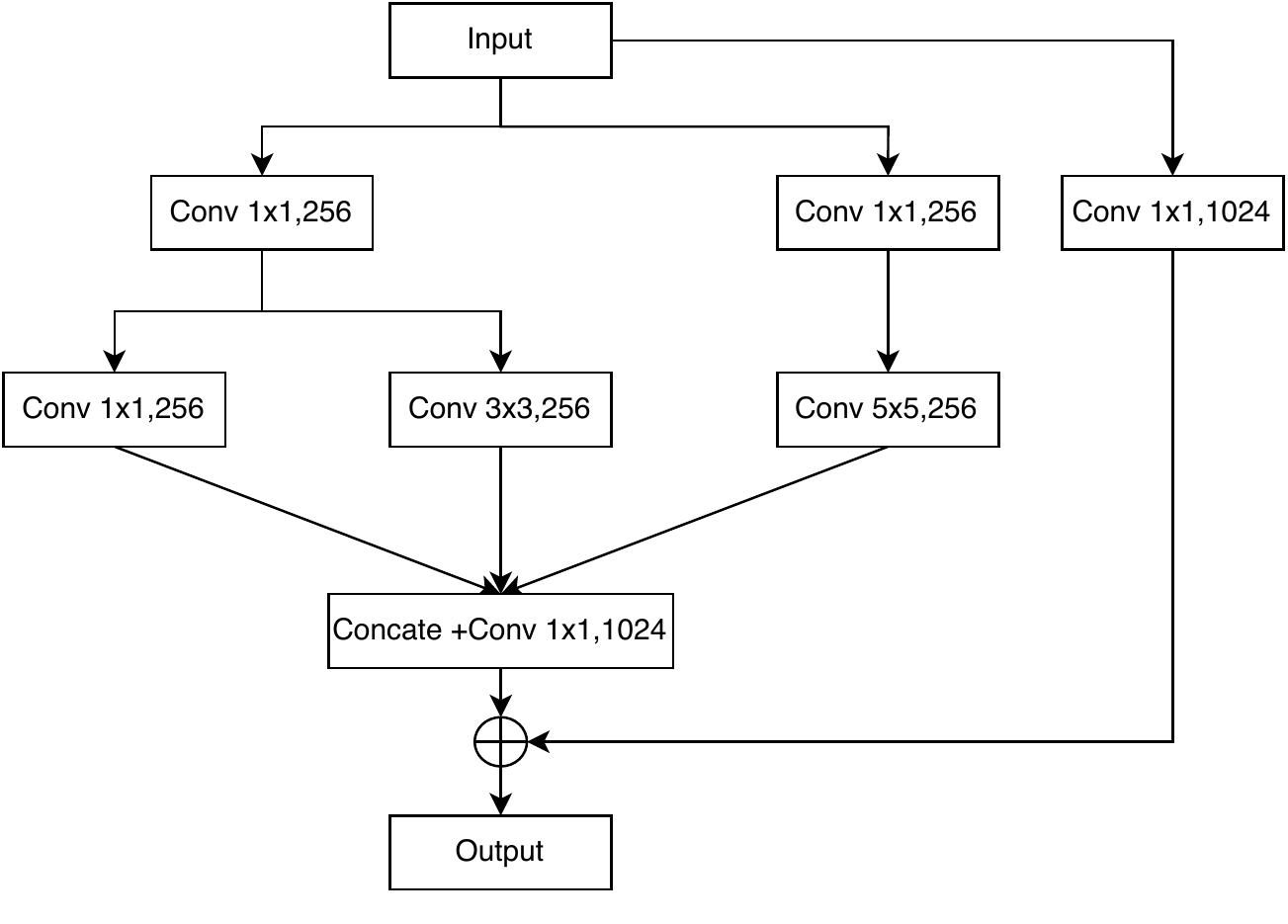}}
    \subfigure[]{\includegraphics[width=0.21\textwidth]{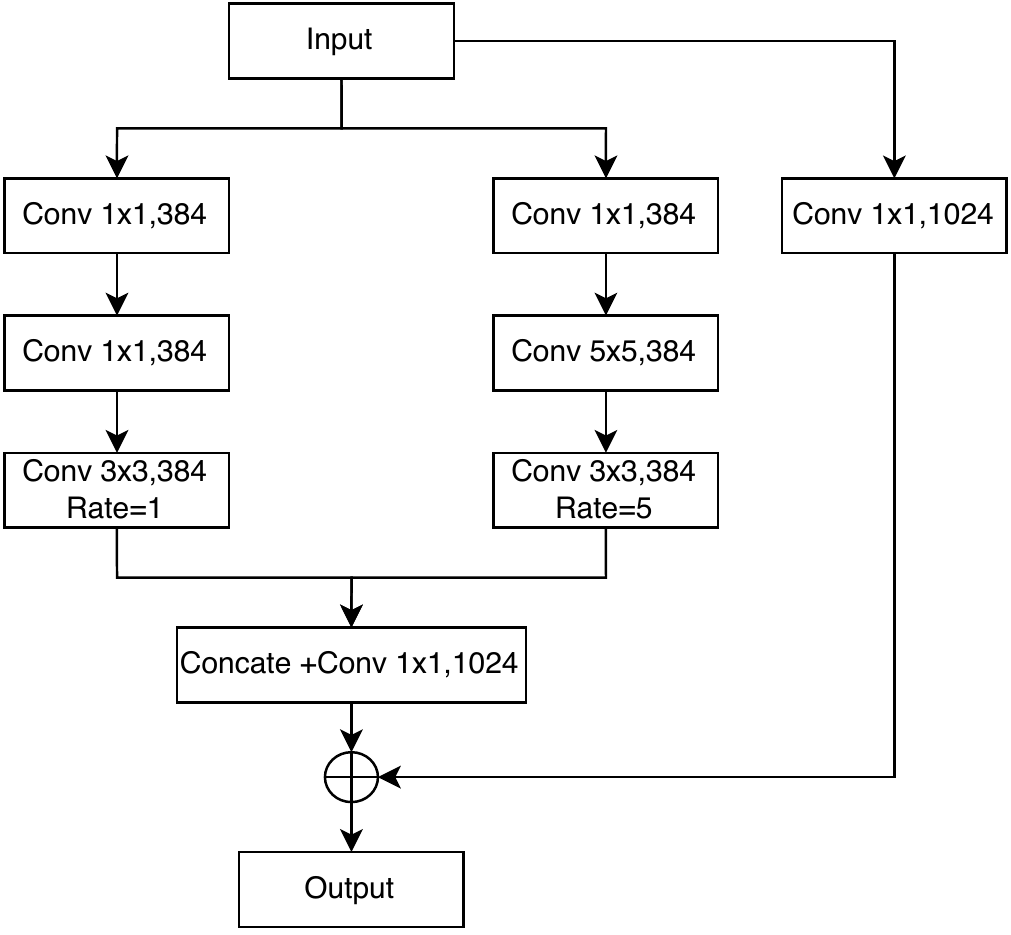}} 
    \subfigure[]{\includegraphics[width=0.21\textwidth]{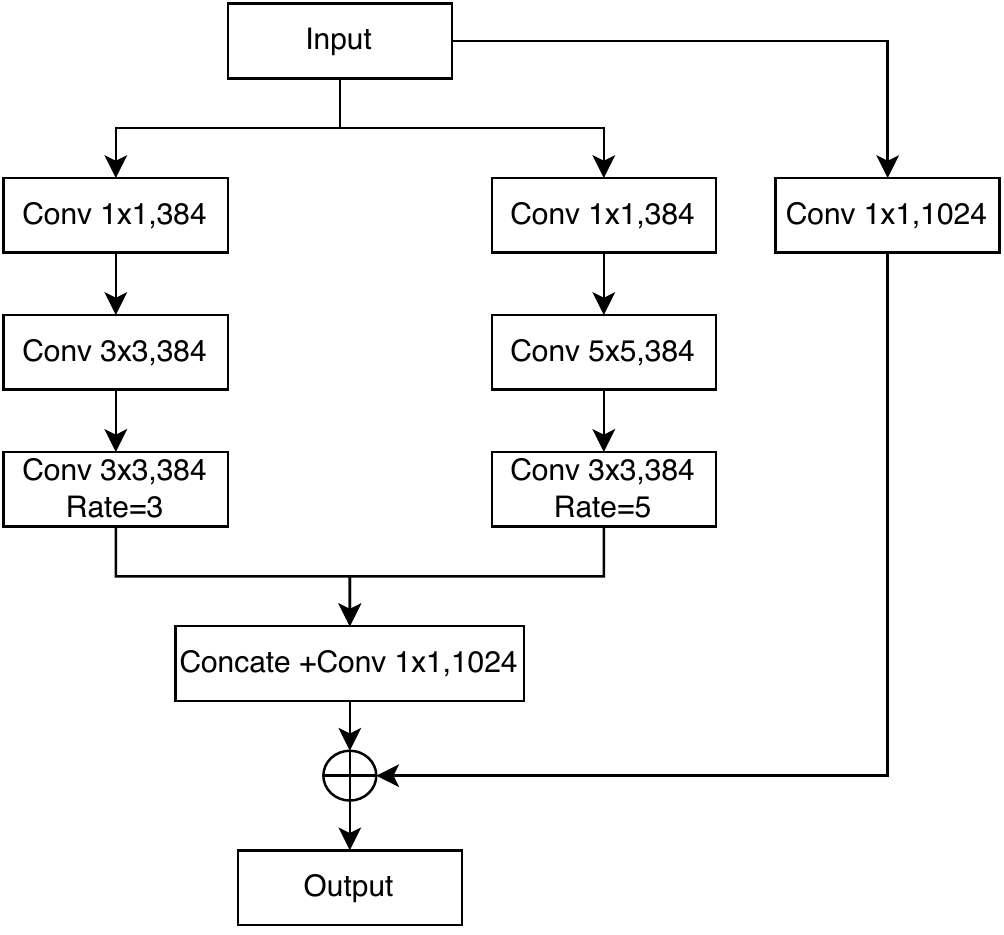}}
    \vspace{-1.3em}
    \caption{\small{(a) The original Multiscale Adapter; (b) Without dilated Convolutions; (c) Without $3\times3$ convolution branch; (d) Without $1\times1$ Convolution Branch. A batch normalization layer and a ReLU layer are cascaded after every convolution operator. Conv$1\times1$,$256$ means using $1\times1$ convolution with $256$ channels.}}
    \vspace{-1.6em}
    \label{fig:foobar}
\end{figure}


In consideration of the limited data scale in the face forgery detection task, we freeze the main parameters (i.e., Transformer layers) of the SAM encoder and insert learnable specific modules to mine task-aware forgery clues. Specifically, we incorporate a concise and efficient  Multiscale Adapter module along each transformer layer to capture forgery clues with diverse receptive fields using a multi-scale fashion. \par

First, we tokenize the input image $x$ into visual tokens $x_{p}=P(x)$ via the Patch Embedding layer (denoted as $P(\cdot)$). During this process, defaulted patch size 14$\times$14 is used. Then $N$ fixed Transformer layers $L_{i},  i\in \left\{1,...,N \right\}$ with learnable Multiscale Adapter $R(\cdot)$ are used for extracting short- and long-range contextual features $Z_{\text{tran}}$, which is then mapped to task-aware forgery features $F$ via a dimension-matched learnable linear Task Head $T(\cdot)$. The feature encoding procedure can be formulated as
\begin{equation}
\vspace{-0.8em}
\begin{split}
Z_{\text{tran}}&=L_{n}(...L_{2}(R(L_{1}(R(x_{p}))), \\
F&=T(Z_{\text{tran}}).
\end{split}
\end{equation}



 As for the Multiscale Adapter (see the middle red block in Fig. \ref{Fig.main1} and Fig. \ref{tab:my-table5}(a)), the output features $x'$ of each Transformer layer are passed by a $1 \times 1$ convolution, and then split into three branches \-$f(\cdot)$, $g(\cdot)$, and $h(\cdot)$. Each branch uses different convolution kernel sizes and dilated rates for complementary forgery context mining. Therefore, the multiscale short- and long-range features $Sout_{1}$ can be formulated as 
\begin{equation}\small
\vspace{-0.3em}
\begin{split}
S\text{out}^{'}&=\text{Concat}(f(\text{Conv}_{1 \times 1}(x')), g(\text{Conv}_{1 \times 1}(x')), h(\text{Conv}^{'}_{1 \times 1}(x'))), \\
S\text{out}_{1}&=\text{Conv}^{''}_{1\times 1}(S\text{out}^{'}),
\end{split}
\vspace{-0.5em}
\end{equation}
where $f(\cdot)$, $g(\cdot)$ and $h(\cdot)$ denote a covolution with kernel size $1 \times 1$ cascaded with a covolution with kernel size $3 \times 3$ and dilated rates $1$, a covolution with kernel size $3 \times 3$ cascaded with a covolution with kernel size $3 \times 3$ and dilated rates $3$, and a covolution with kernel size $5 \times 5$ cascaded with a covolution with kernel size $3 \times 3$ and dilated rates $5$, respectively. $S\text{out}^{'}$ is the result of merging features of the three branches, which is then re-projected to the original channel size via a $1 \times 1$ convolution to obtain $S\text{out}_{1}$.

Finally, the resultant multi-scale features $S\text{out}_{1}$ are added together to the original features $x'$ passed over a  $1\times 1$ convolution operation $\text{Conv}^{'''}_{1\times 1}$, ensuring the preservation of the original information. The final multiscale contextual features $S\text{out}$ can be formulated as
\begin{center}
\vspace{-1.8em}
\begin{equation}
S\text{out}=S\text{out}_{1}+\text{Conv}^{'''}_{1 \times 1}(x').
\end{equation}
\end{center}

In terms of the structure of Multiscale Adapter, as shown in Fig. \ref{fig:foobar}, three kinds of variants (i.e., Multiscale Adapter-B, Multiscale Adapter-C, Multiscale Adapter-D) are also investigated according to different scalabilities of receptive fields. The ablation study among them can be found in Table \ref{tab:my-table5}.


\vspace{-0.8em}
\subsection{Reconstruction Guided Attention}
\vspace{-0.3em}
In order to enhance the sensitivity to deep forged regions and explore the common and compact feature patterns of real faces, we propose a reconstruction learning method, namely Reconstruction Guided Attention (RGA). 



 In the training process, we simulate the forged faces by introducing white noise $G(\cdot)$ on the real faces. Based on the noisy inputs, the model gradually performs feature reconstruction to obtain the reconstructed features $F^{Gau}$. 
\begin{equation}
\begin{split}
x^{\text{Gau}}&=G\left(x\right)\:,\\
F^{\text{Gau}}&=\Phi\left(x^{\text{Gau}}\right),
\end{split}
\end{equation}
where $\Phi$ denotes the whole image encoder. After the feature reconstruction process performed by the image encoder, we compute the absolute difference $S$ between the original features and the reconstructed features. In this calculation, the function $|.|$ represents the absolute value function.
\begin{center}
\vspace{-0.8em}
\begin{equation}
S=\left|F^{\text{Gau}}-F\right|.
\end{equation}
\end{center}
Subsequently, an enhancer $\varphi(\cdot)$ with $1\times1$ convolution is employed to highlight and enhance regions that might contain forgeries, which is cascaded with the Softmax function layer $\alpha(\cdot)$ to generate the forgery-aware attention map. Finally, we perform element-wise multiplication based refinement operation $\otimes$ between the obtained attention weights and the original features to obtain the final features $F_{\text{final}}$, which are then sent for the mask decoder. The procedure can be formulated as:
\begin{equation}
\vspace{-0.4em}
F_{\text{final}}=\left[ \alpha(\varphi(S)) \otimes  \varphi(F)\right ]+F.
\end{equation}

After obtaining the features $F$ from the real faces and the reconstructed features $F^{\text{Gau}}$ from the anomalies/forgeries simulation, we calculate the reconstruction loss $\mathcal{L}_{\text{rec}}$ for each batch $M$ with $L1$ norm. Notably, the reconstruction loss $L_{\text{rec}}$ is exclusively trained on real samples. Ablation studies on calculating $\mathcal{L}_{\text{rec}}$ for fake faces and all (real+fake) faces can be found in Table \ref{tab:my-table6}. 
\begin{equation}
\vspace{-0.4em}
\mathcal{L}_{\text{rec}}=\frac{1}{M} \sum_{i \in M}\left\|F_i^{\text{Gau}}-F_i\right\|_1. 
\end{equation}

In the training stage, the RGA module leverages the abnormal simulated faces as one of the inputs and gradually recovers the intrinsic features of the real faces. Through this reconstruction process, SAM models can better understand the common and compact feature patterns of real faces, and even pay more attention to unknown forged regions in the inference stage. 

\vspace{-1.3em}
\subsection{Loss Function}
\vspace{-0.5em}
The overall loss function $\mathcal{L}_{\text {overall}}$ of DADF consists of three components: segmentation loss $\mathcal{L}_{\text{seg}}$, classification loss $\mathcal{L}_{\text{cls}}$, and feature reconstruction loss $ \mathcal{L}_{\text{rec}}$. The segmentation loss $\mathcal{L}_{\text {seg }}$ represents the semantic loss, while the binary cross-entropy loss $\mathcal{L}_{\text {cls}}$ measures the binary real/fake classification error. The feature reconstruction los $\mathcal{L}_{\text{rec}}$ captures the reconstruction error.
\begin{equation}
\vspace{-0.8em}
\mathcal{L}_{\text {overall}}=\mathcal{L}_{\text{seg }}+\lambda_1 \mathcal{L}_{\text{rec}}+\lambda_{2} \mathcal{L}_{\text {cls}},
\label{qa}
\end{equation}
where the hyperparameters $\lambda_1$
 and $\lambda_2$
 are used to balance the different components of the loss, which are set to 0.1 according to empirical observations.

 \vspace{-1.3em}
\section{Experiments}
\vspace{-0.8em}
\subsection{Datasets and Performance Metrics}
\vspace{-0.5em}
\noindent\textbf{Datasets.}   The \textbf{FaceForensics++} (FF++) \cite{rossler2018faceforensics} utilizes four different algorithms: Deepfakes (DF), Face2Face (F2F), FaceSwap (FS) and NeuralTextures (NT) to generate forgery faces. The video data also provide versions with different compression ratios: original quality (quantization = 0), high quality (HQ, quantization = 23) and low quality (LQ, quantization = 40). 
The \textbf{DF-TIMI} \cite{korshunov2018deepfakes} dataset contained 16 pairs of similar people, each of whom lived 10 videos. It includes 960 videos: 320 genuine and 640 forged using FaceSwap technology. Among the forged videos, 320 are high quality (HQ) and 320 are low quality (LQ).
The \textbf{DFD} \cite{rossler2018faceforensics} was created specifically for DeepFake technology. These videos were procured from the YouTube platform, consisting of 363 authentic videos and 3068 fabricated videos, which have been further categorized as high-quality (HQ) and low-quality (LQ).
The \textbf{FMLD} \cite{kong2022detect} comprises 40,000 synthetic faces generated using StyleGAN and PGGAN models, and an additional 40,000 attribute-manipulated faces using StarGAN and AttGAN. Specifically, attribute manipulations include glasses and hair regions.


\vspace{0.3em}
\noindent\textbf{Evaluation metrics.}
Two commonly used metrics, namely Binary Classification Accuracy (PBCA) and Inverse Intersection Non-Containment (IINC) \cite{dang2020detection} are employed for forgery localization. For fair comparisons, we follow the same evaluation protocols as \cite{kong2022detect} for face forgery localization. In terms of evaluating the performance of face forgery detection, Accuracy (ACC) is adopted.

\begin{table}[t]
\centering
\caption{\small{Results of the forgery localization on FF++ (HQ) \cite{rossler2018faceforensics} and FMLD \cite{kong2022detect}.}}
\label{tab:my-table1}
\resizebox{0.55\textwidth}{!}{\begin{tabular}{c|cc|cc}
\hline
Dataset & \multicolumn{2}{c|}{FF++ (HQ)} & \multicolumn{2}{c}{FMLD} \\ \hline
Methods & \multicolumn{1}{c}{PBCA(\%)}$\uparrow$ \quad & IINC(\%)$\downarrow$ & \multicolumn{1}{c}{PBCA(\%)}$\uparrow$ \quad & IINC(\%)$\downarrow$ \\ \hline
Multitask \cite{nguyen2019multi}& \multicolumn{1}{c}{94.88} & 4.46 & \multicolumn{1}{c}{98.59} & 3.52 \\ \hline
DFFD Reg \cite{dang2020detection}& \multicolumn{1}{c}{94.85} & 4.57 & \multicolumn{1}{c}{98.72} & 3.31 \\ \hline
DFFD Mam \cite{dang2020detection}& \multicolumn{1}{c}{91.45} & 13.09 & \multicolumn{1}{c}{96.86} & 23.93 \\ \hline
Locate \cite{kong2022detect}& \multicolumn{1}{c}{95.77} & 3.62 & \multicolumn{1}{c}{99.06} & \textbf{2.53} \\ \hline
SAM \cite{kirillov2023segment}& \multicolumn{1}{c}{92.97} & 4.25 & \multicolumn{1}{c}{97.29} & 3.40 \\ \hline
SAM+LoRA \cite{zhou2014low}& \multicolumn{1}{c}{93.12} & 4.78 & \multicolumn{1}{c}{98.06} & 3.51 \\ \hline
SAM+Prompt \cite{qiu2023learnable}& \multicolumn{1}{c}{94.60} & 3.48 & \multicolumn{1}{c}{98.01} & 2.82 \\ \hline
\textbf{DADF (Ours)} & \multicolumn{1}{c}{\textbf{96.64}} & \textbf{3.21} & \multicolumn{1}{c}{\textbf{99.26}} & 2.64 \\ \hline
\end{tabular}}
\vspace{-0.8em}
\end{table}

\vspace{-1.3em}
\subsection{Implementation Details}
\vspace{-0.4em}
We use the SAM-based \cite{kirillov2023segment} ViT-H model as the backbone with a null input prompt setting. We train models with a batch size of 4 and adopt the AdamW optimizer. We employ the cosine decay method. The initial learning rate is set to 0.05. For the FF++ dataset, we conduct 30 epochs of training, while the FMLD dataset requires 50 epochs to train the model effectively.To adapt the learning rate, a step learning rate scheduler is employed. As for the RGA module, white noise is employed as a noise source, which is incorporated into the data using a normal distribution with a zero mean and a variance of 1e-6.

 

\vspace{-1.3em}
\subsection{Intra-dataset Testing}
\vspace{-0.5em}

\begin{table}[t]
\centering
\caption{\small{The forgery detection performance (ACC(\%)) on FF++ (LQ) \cite{rossler2018faceforensics}.}}
\label{tab:my-table2}
\resizebox{0.52\textwidth}{!}{\begin{tabular}{c|c|c|c|c|c}
\hline
Methods & DF \quad & FF \quad & FS \quad & NT \quad & Average \\ \hline
Steg. Features\cite{fridrich2012rich}& 67.00 & 48.00 & 49.00 & 56.00 & 55.00 \\ \hline
Cozzolino \cite{cao2022end}& 75.00 & 56.00 & 51.00 & 62.00 & 61.00 \\ \hline
Bayar \& Stamm  \cite{bayar2016deep}& 87.00 & 82.00 & 74.00 & 74.00 & 79.25 \\ \hline
Rahmouni \cite{afchar2018mesonet}& 80.00 & 62.00 & 59.00 & 59.00 & 65.00 \\ \hline
MesoNet \cite{liu2021p}& 90.00 & 83.00 & 83.00 & 75.00 & 82.75 \\ \hline
SPSL \cite{chollet2017xception}& 93.48 & 86.02 & 92.26 & \textbf{92.26} & 91.00 \\ \hline
Xception \cite{chollet2017xception}& 97.16 & 91.02 & 96.71 & 82.88 & 91.94 \\ \hline
Locate \cite{kong2022detect}& 97.25 & 94.46 & 97.13 & 84.63 & 93.36 \\ \hline
SAM \cite{kirillov2023segment}& { 89.32} & { 84.56} & { 91.19} & { 80.01} & 86.27 \\ \hline
SAM+LoRA \cite{zhou2014low}& { 90.12} & { 85.41} & { 91.28} & { 80.15} & 86.74 \\ \hline
SAM+Prompt \cite{qiu2023learnable}& { 97.34} & { 95.84} & { 97.44} & { 84.72} & 93.83 \\ \hline
\textbf{DADF (Ours)} & { \textbf{99.02} }& { \textbf{98.92} }& { \textbf{98.23}} & { 87.61} & \textbf{95.94 }\\ \hline
\end{tabular}}
\vspace{-1.3em}
\end{table}

\begin{table}[t]
\caption{\small{Results of cross-dataset face forgery detection.}}
\centering
\label{tab:my-table3}
\resizebox{0.9\textwidth}{!}{
\begin{tabular}{c|cccccccc}
\hline
Dataset & \multicolumn{2}{c|}{DFD (LQ)} & \multicolumn{2}{c|}{DF-TIMIT (HQ)} & \multicolumn{2}{c|}{DF-TIMIT (LQ)} & \multicolumn{2}{c}{Average} \\ \hline
Method & \multicolumn{1}{c}{AUC(\%)}$\uparrow$ \quad & \multicolumn{1}{c|}{EER(\%)$\downarrow$}  & \multicolumn{1}{c}{AUC(\%)}$\uparrow$ \quad & \multicolumn{1}{c|}{EER(\%)$\downarrow$} & \multicolumn{1}{c}{AUC(\%)}$\uparrow$ \quad & \multicolumn{1}{c|}{EER(\%)$\downarrow$} & \multicolumn{1}{c}{AUC(\%)}$\uparrow$ \quad & EER(\%)$\downarrow$ \\ \hline
MesoNet \cite{afchar2018mesonet}& \multicolumn{1}{c}{52.25} & \multicolumn{1}{c|}{48.65} & \multicolumn{1}{c}{33.61} & \multicolumn{1}{c|}{60.16} & \multicolumn{1}{c}{45.08} & \multicolumn{1}{c|}{53.04} & \multicolumn{1}{c}{34.64} & 53.95 \\ \hline
MesoIncep4 \cite{afchar2018mesonet}& \multicolumn{1}{c}{\textbf{63.27}} & \multicolumn{1}{c|}{40.37} & \multicolumn{1}{c}{16.12} & \multicolumn{1}{c|}{76.18} & \multicolumn{1}{c}{27.47} & \multicolumn{1}{c|}{66.77} & \multicolumn{1}{c}{35.62} & 61.10 \\ \hline
ResNet50 \cite{he2016deep}& \multicolumn{1}{c}{60.61} & \multicolumn{1}{c|}{42.23} & \multicolumn{1}{c}{41.95} & \multicolumn{1}{c|}{55.97} & \multicolumn{1}{c}{47.27} & \multicolumn{1}{c|}{52.33} & \multicolumn{1}{c}{49.94} & 50.17 \\ \hline
Face X-ray \cite{li2020face}& \multicolumn{1}{c}{62.89} & \multicolumn{1}{c|}{39.58} & \multicolumn{1}{c}{42.52} & \multicolumn{1}{c|}{55.07} & \multicolumn{1}{c}{50.05} & \multicolumn{1}{c|}{\textbf{49.11}} & \multicolumn{1}{c}{51.81} & 47.92 \\ \hline
DFFD \cite{dang2020detection}& \multicolumn{1}{c}{60.60} & \multicolumn{1}{c|}{42.32} & \multicolumn{1}{c}{32.91} & \multicolumn{1}{c|}{61.16} & \multicolumn{1}{c}{39.32} & \multicolumn{1}{c|}{57.06} & \multicolumn{1}{c}{44.27} & 53.51 \\ \hline
Multi-task \cite{nguyen2019multi}& \multicolumn{1}{c}{58.61} & \multicolumn{1}{c|}{44.49} & \multicolumn{1}{c}{16.53} & \multicolumn{1}{c|}{77.86} & \multicolumn{1}{c}{15.59} & \multicolumn{1}{c|}{78.50} & \multicolumn{1}{c}{30.24} & 66.95 \\ \hline
F3Net \cite{qian2020thinking}& \multicolumn{1}{c}{58.89} & \multicolumn{1}{c|}{39.87} & \multicolumn{1}{c}{29.12} & \multicolumn{1}{c|}{58.33} & \multicolumn{1}{c}{45.67} & \multicolumn{1}{c|}{52.72} & \multicolumn{1}{c}{44.56} & 50.30 \\ \hline
Xception \cite{chollet2017xception}& \multicolumn{1}{c}{59.73} & \multicolumn{1}{c|}{43.12} & \multicolumn{1}{c}{33.82} & \multicolumn{1}{c|}{62.83} & \multicolumn{1}{c}{40.79} & \multicolumn{1}{c|}{57.44} & \multicolumn{1}{c}{44.78} & 54.46 \\ \hline
SAM \cite{kirillov2023segment}& \multicolumn{1}{c}{50.61} & \multicolumn{1}{c|}{49.13} & \multicolumn{1}{c}{43.19} & \multicolumn{1}{c|}{57.94} & \multicolumn{1}{c}{45.71} & \multicolumn{1}{c|}{54.39} & \multicolumn{1}{c}{46.50} & 53.82 \\ \hline
SAM+LoRA \cite{zhou2014low}& \multicolumn{1}{c}{53.71} & \multicolumn{1}{c|}{48.29} & \multicolumn{1}{c}{43.64} & \multicolumn{1}{c|}{56.67} & \multicolumn{1}{c}{47.64} & \multicolumn{1}{c|}{53.02} & \multicolumn{1}{c}{48.33} & 52.66 \\ \hline
SAM+Prompt \cite{qiu2023learnable}& \multicolumn{1}{c}{57.25} & \multicolumn{1}{c|}{45.28} & \multicolumn{1}{c}{44.32} & \multicolumn{1}{c|}{55.07} & \multicolumn{1}{c}{48.17} & \multicolumn{1}{c|}{52.54} & \multicolumn{1}{c}{49.91} & 50.96 \\ \hline
\textbf{DADF (Ours)} & \multicolumn{1}{c}{ 63.21} & \multicolumn{1}{c|}{\textbf{39.52}} & \multicolumn{1}{c}{\textbf{46.37}} & \multicolumn{1}{c|}{\textbf{53.20}} & \multicolumn{1}{c}{\textbf{50.62}} & \multicolumn{1}{c|}{49.74} & \multicolumn{1}{c}{\textbf{53.40}} & \textbf{47.48} \\ \hline
\end{tabular}}
\vspace{-0.8em}
\end{table}

\begin{table}[t]
\centering
\caption{\small{Ablation studies on the FF++ (HQ) \cite{rossler2018faceforensics} dataset.}}
\label{tab:my-table4}
\resizebox{0.6\textwidth}{!}{\begin{tabular}{c|c|c|cc|c}
\hline
\multirow{2}{*}{Baseline (SAM)} & \multirow{2}{*}{Multiscale Adapter} & \multirow{2}{*}{RGA} & \multicolumn{2}{c|}{Localization} & Detection \\ \cline{4-6}
 &  &  & \multicolumn{1}{c}{PBLA(\%)$\uparrow$} \quad & IINC(\%)$\downarrow$ & ACC(\%)$\uparrow$ \\ \hline
\checkmark &  &  & \multicolumn{1}{c}{92.97} & 4.25 & 86.27 \\ \hline
\checkmark & \checkmark &  & \multicolumn{1}{c}{96.31} & 3.40 & 94.48 \\ \hline
\checkmark &  & \checkmark & \multicolumn{1}{c}{95.61} & 3.96 & 92.64 \\ \hline
\checkmark & \checkmark & \checkmark & \multicolumn{1}{c}{\textbf{96.64}} & \textbf{3.21} & \textbf{95.94} \\ \hline
\end{tabular}}
\vspace{-1.6em}
\end{table}

 

\begin{table}[]
\centering
\caption{\small{Ablations of different Multiscale Adapters on FF++ (HQ) \cite{rossler2018faceforensics} dataset.}}
\label{tab:my-table5}
\resizebox{0.49\textwidth}{!}{\begin{tabular}{c|cc|c}
\hline
\multirow{2}{*}{Module} & \multicolumn{2}{c|}{Localization} & Detection \\ \cline{2-4} 
 & \multicolumn{1}{c}{PBLA(\%)$\uparrow$} \quad & IINC(\%)$\downarrow$ \quad & ACC(\%)$\uparrow$ \\ \hline
Multiscale Adapter-B & \multicolumn{1}{c}{96.52} & 3.43 & 95.79 \\ \hline
Multiscale Adapter-C & \multicolumn{1}{c}{96.57} & 3.31 & 95.87 \\ \hline
Multiscale Adapter-D & \multicolumn{1}{c}{96.61} & 3.26 & 95.91 \\ \hline
Multiscale Adapte & \multicolumn{1}{c}{\textbf{96.64}} & \textbf{3.21} & \textbf{95.94 }\\ \hline
\end{tabular}}
\vspace{-0.8em}
\end{table}

\noindent\textbf{Results of face forgery localization.}
Table \ref{tab:my-table1} presents the results of the forgery localization on FF++ (HQ) \cite{rossler2018faceforensics} and FMLD \cite{kong2022detect}. The proposed DADF outperform the classical face forgery localization model \cite{kong2022detect} by 0.87\% and 0.2\% PBCA on FF++ (HQ) and FMLD, respectively. We can also find from the results of SAM \cite{kirillov2023segment} that direct finetuning SAM cannot achieve acceptable face forgery localization performance due to its heavy model parameters and limited task-aware data. Despite slight improvement via parameter-efficient fine-tuning strategies, SAM with LoRA or Prompt still has performance gaps with the previous localization method \cite{kong2022detect}. Thanks to the rich forgery contexts from the Multiscale Adapter and the strong forgery attention ability of RGA module, the proposed DADF improves baseline SAM \cite{kirillov2023segment} by 3.67\%/-1.04\% and 1.97\%/-0.76\% PBCA/IINC on FF++ (HQ) and FMLD, respectively.

\vspace{0.3em}
\noindent\textbf{Results of face forgery detection.}
Table \ref{tab:my-table2} presents the detection accuracy (ACC) of our model on various forgery techniques, namely Deepfake (DF), Face2Face (FF), FaceSwap (FS), and NeuralTextures (NT), using the challenging FF++ (LQ) \cite{rossler2018faceforensics}. 
It is clear that the proposed DADF performs significant improvements in classification accuracy compared to previous methods among different forgery techniques. This highlights the effectiveness of our Multiscale Adapter and RGA module in enhancing the detection capabilities, compared with the original SAM \cite{kirillov2023segment} and its variants (SAM+LoRA \cite{zhou2014low} and SAM+Prompt \cite{qiu2023learnable}). Specifically, the proposed DADF improves more than 3\% ACC compared with the second-best method on Face2Face detection.

\vspace{-1.3em}
\subsection{Cross-dataset Testing}
\vspace{-0.5em}

In order to assess the generalization ability of our method on unseen domains and unknown deepfakes, we conducted cross-dataset experiments by training and testing on different datasets. Specifically, we train models on FF++ (LQ), and then test them on DFD (LQ), DF-TIMIT (HQ), and DF-TIMIT (LQ). The results shown in Table \ref{tab:my-table3} demonstrate that the proposed DADF outperforms other methods in terms of average performance among the three testing settings. Compared with SAM and its variants, the significant improvement of DADF in performance is attributed to 1) the introduction of the Multiscale Adapter, which enables forgery feature learning from diverse receptive fields; and 2) the attentional forgery feature refinement via the RGA module, enhancing the robustness under domain shifts and perception of forgery regions.

\vspace{-1.3em}
\subsection{Ablation Study}
\vspace{-0.5em}
To validate the effectiveness of the Multiscale Adapter and Reconstruction Guided Attention module, ablation experiments are conducted on FF++ (HQ).

\vspace{0.3em}
\noindent\textbf{Efficacy of the Multiscale Adapter.}
It can be seen from the first two rows of Table \ref{tab:my-table4} that compared with baseline SAM-only fine-tuning, SAM with Multiscale Adapter improves 3.34\%/-0.85\% PBLA/IINC for forgery localization and 8.21\% ACC for forgery detection on the FF++ (HQ). Considering different configurations (see Fig. \ref{fig:foobar}) of Multiscale Adapter, Table \ref{tab:my-table5} demonstrates that removing dilated convolutions results in the largest accuracy drop, while the best performance is achieved by incorporating multiscale convolution modules alongside dilated convolutions.

\begin{table}[t]
\centering
\caption{\small{Ablation studies of $\mathcal{L}_{\text{rec}}$ calculation on FF++ (HQ) \cite{rossler2018faceforensics} dataset.}}
\label{tab:my-table6}
\resizebox{0.42\textwidth}{!}{\begin{tabular}{c|cc|c}
\hline
\multirow{2}{*}{Data} & \multicolumn{2}{c|}{Localization} & Detection \\ \cline{2-4} 
  & \multicolumn{1}{c}{PBLA(\%)$\uparrow$} \quad & IINC(\%)$\downarrow$ & ACC(\%)$\uparrow$ \\ \hline
Real \& Fake & \multicolumn{1}{c}{92.15} & 3.62 & 91.42 \\ \hline
Fake &  \multicolumn{1}{c}{93.34} & 3.56 & 92.16 \\ \hline
Real &  \multicolumn{1}{c}{\textbf{96.64}} & \textbf{3.21} & \textbf{95.94} \\ \hline
\end{tabular}}
\vspace{-0.8em}
\end{table}

\vspace{0.3em}
\noindent\textbf{Efficacy of the RGA.}
As shown in the last two rows of Table \ref{tab:my-table4}, equipping with RGA module can improve the baseline SAM by 2.64\%/-0.29\% PBLA/IINC for forgery localization and 6.37\% ACC for forgery detection on the FF++ (HQ). Similarly, based on the SAM with Multiscale Adapter, the RGA module can further benefit the forgery localization by 0.33\% PBLA and detection by 1.46\% ACC. As for the loss function $\mathcal{L}_{\text{rec}}$ calculation for RGA, it can be seen from Table \ref{tab:my-table6} that the performance drops sharply when $\mathcal{L}_{\text{rec}}$ calculated for fake faces and all (real+fake) faces, which might result from the redundant features of real faces and less attention on anomalies.

\vspace{-1.3em}
\subsection{Visualization and Discussion}
\vspace{-0.5em}
\begin{figure}[t] 
\centering 
\includegraphics[width=0.9\textwidth]{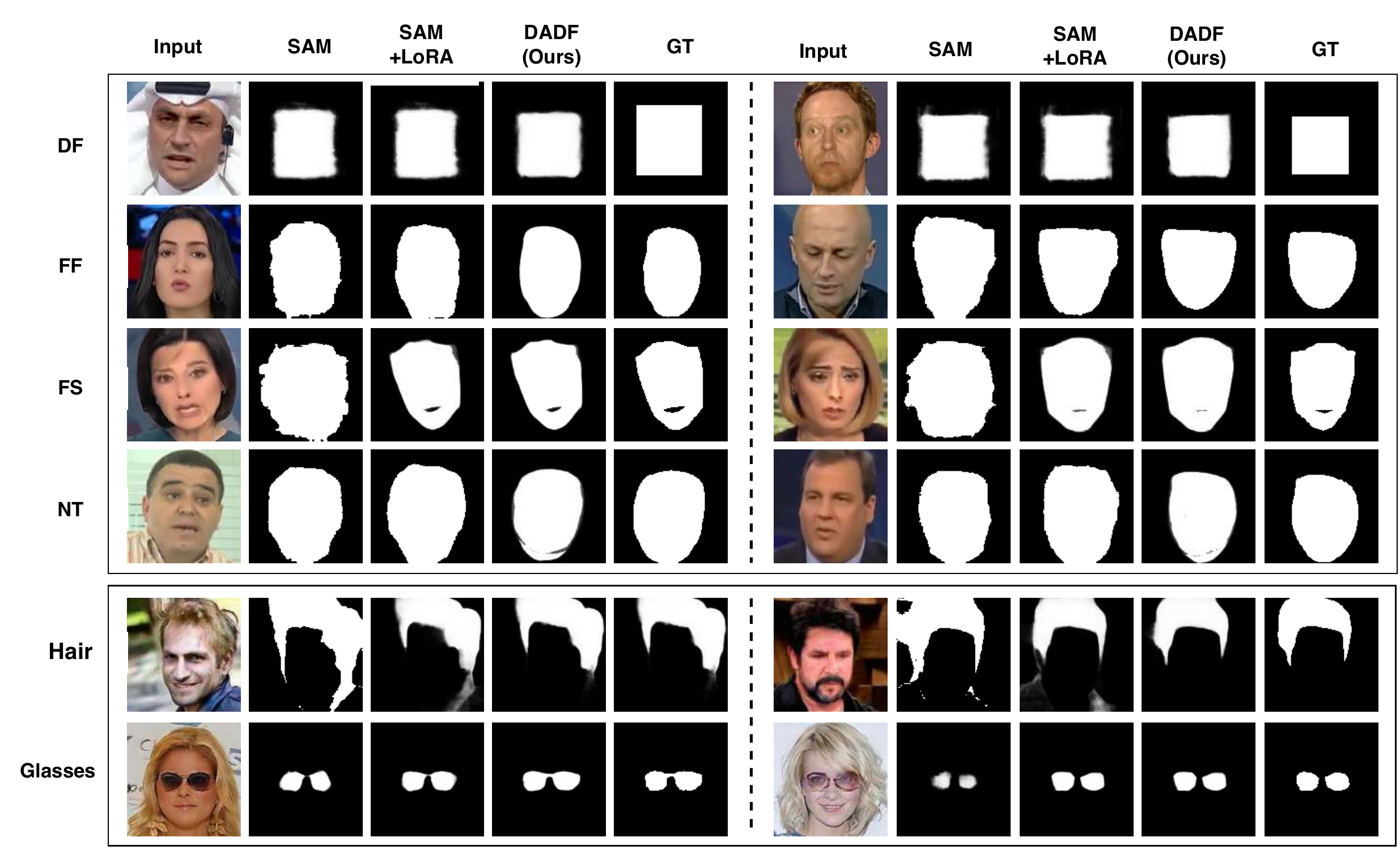} 
\vspace{-1.3em}
\caption{Visualization of face forgery localization results of various methods on the FF++ (HQ) dataset (DF, FF, FS, and NT) \cite{rossler2018faceforensics} and FMLD (Hair and Glasses) \cite{kong2022detect}. } 
\label{Fig.main2} 
\vspace{-1.3em}
\end{figure}

We visualize some representative forgery samples and their mask labels as well as predictions in Fig. \ref{Fig.main2}. It is evident that the forgery localization quality from the proposed DADF outperforms SAM and its LoRA fine-tuning in accurately localizing and closely resembling the ground truth, particularly in fine-grained details such as edge, boundary, and face-head contexts.



Besides, the proposed Multiscale Adapter is a parameter-efficient fine-tuning strategy alternative to tune the entire Transformer layers. Remarkably, by adjusting only 18.64\% parameters of the SAM, substantial benefits on face forgery detection and localization are achieved, including reduced training costs and improved practical performance.

%
%
%

\vspace{-0.8em}
\section{Conclusion}
\vspace{-0.5em}
In this paper, we introduce a Segment Anything Model based face forgery detection and localization framework, namely Detect Any Deepfakes (DADF). Specifically, we propose the Multiscale Adapter and Reconstruction Guided Attention (RGA) to efficiently fine-tune SAM with rich contextual forgery clues and enhance the robustness of forgery localization. Extensive experimental results validate the effectiveness of the proposed DADF across different qualities of face images and even under cross-domain scenarios.

\bibliographystyle{splncs04}
\bibliography{ref}

\begin{thebibliography}{10}
\providecommand{\url}[1]{\texttt{#1}}
\providecommand{\urlprefix}{URL }
\providecommand{\doi}[1]{https://doi.org/#1}

\bibitem{afchar2018mesonet}
Afchar, D., Nozick, V., Yamagishi, J., Echizen, I.: Mesonet: a compact facial
  video forgery detection network. In: IEEE WIFS (2018)

\bibitem{agarwal2021detecting}
Agarwal, S., Farid, H.: Detecting deep-fake videos from aural and oral
  dynamics. In: IEEE CVPR (2021)

\bibitem{bappy2017exploiting}
Bappy, J.H., Roy-Chowdhury, A.K., Bunk, J., Nataraj, L., Manjunath, B.:
  Exploiting spatial structure for localizing manipulated image regions. In:
  IEEE ICCV (2017)

\bibitem{bayar2016deep}
Bayar, B., Stamm, M.C.: A deep learning approach to universal image
  manipulation detection using a new convolutional layer. In: ACM workshop on
  information hiding and multimedia security (2016)

\bibitem{cao2022end}
Cao, J., Ma, C., Yao, T., Chen, S., Ding, S., Yang, X.: End-to-end
  reconstruction-classification learning for face forgery detection. In: CVPR
  (2022)

\bibitem{chen2023sam}
Chen, T., Zhu, L., Ding, C., Cao, R., Zhang, S., Wang, Y., Li, Z., Sun, L.,
  Mao, P., Zang, Y.: Sam fails to segment anything? -- sam-adapter: Adapting
  sam in underperformed scenes: Camouflage, shadow, and more (2023)

\bibitem{chierchia2011prnu}
Chierchia, G., Parrilli, S., Poggi, G., Verdoliva, L., Sansone, C.: Prnu-based
  detection of small-size image forgeries. In: IEEE DSP (2011)

\bibitem{chollet2017xception}
Chollet, F.: Xception: Deep learning with depthwise separable convolutions. In:
  IEEE CVPR (2017)

\bibitem{clark2020electra}
Clark, K., Luong, M.T., Le, Q.V., Manning, C.D.: Electra: Pre-training text
  encoders as discriminators rather than generators. arXiv preprint
  arXiv:2003.10555  (2020)

\bibitem{dang2020detection}
Dang, H., Liu, F., Stehouwer, J., Liu, X., Jain, A.K.: On the detection of
  digital face manipulation. In: IEEE CVPR (2020)

\bibitem{devlin2018bert}
Devlin, J., Chang, M.W., Lee, K., Toutanova, K.: Bert: Pre-training of deep
  bidirectional transformers for language understanding. arXiv preprint
  arXiv:1810.04805  (2018)

\bibitem{fridrich2012rich}
Fridrich, J., Kodovsky, J.: Rich models for steganalysis of digital images.
  IEEE TIFS  (2012)

\bibitem{guan2023collaborative}
Guan, W., Wang, W., Dong, J., Peng, B., Tan, T.: Collaborative feature learning
  for fine-grained facial forgery detection and segmentation. arXiv preprint
  arXiv:2304.08078  (2023)

\bibitem{guo2023hierarchical}
Guo, X., Liu, X., Ren, Z., Grosz, S., Masi, I., Liu, X.: Hierarchical
  fine-grained image forgery detection and localization. In: IEEE CVPR (2023)

\bibitem{he2016deep}
He, K., Zhang, X., Ren, S., Sun, J.: Deep residual learning for image
  recognition. In: IEEE CVPR (2016)

\bibitem{hochreiter1997long}
Hochreiter, S., Schmidhuber, J.: Long short-term memory. Neural computation
  (1997)

\bibitem{huang2022fakelocator}
Huang, Y., Juefei-Xu, F., Guo, Q., Liu, Y., Pu, G.: Fakelocator: Robust
  localization of gan-based face manipulations. IEEE TIFS  (2022)

\bibitem{kirillov2023segment}
Kirillov, A., Mintun, E., Ravi, N., Mao, H., Rolland, C., Gustafson, L., Xiao,
  T., Whitehead, S., Berg, A.C., Lo, W.Y., et~al.: Segment anything. arXiv
  preprint arXiv:2304.02643  (2023)

\bibitem{kong2022detect}
Kong, C., Chen, B., Li, H., Wang, S., Rocha, A., Kwong, S.: Detect and locate:
  Exposing face manipulation by semantic-and noise-level telltales. IEEE TIFS
  (2022)

\bibitem{korshunov2018deepfakes}
Korshunov, P., Marcel, S.: Deepfakes: a new threat to face recognition?
  assessment and detection. arXiv preprint arXiv:1812.08685  (2018)

\bibitem{li2020face}
Li, L., Bao, J., Zhang, T., Yang, H., Chen, D., Wen, F., Guo, B.: Face x-ray
  for more general face forgery detection. In: IEEE CVPR (2020)

\bibitem{liu2021p}
Liu, X., Ji, K., Fu, Y., Tam, W.L., Du, Z., Yang, Z., Tang, J.: P-tuning v2:
  Prompt tuning can be comparable to fine-tuning universally across scales and
  tasks. arXiv preprint arXiv:2110.07602  (2021)

\bibitem{liy2018exposingaicreated}
LIY, C.M., InIctuOculi, L.: Exposingaicreated fakevideosbydetectingeyeblinking.
  In: IEEE WIFS (2018)

\bibitem{1}
Luk{\'a}{\v{s}}, J., Fridrich, J., Goljan, M.: Detecting digital image
  forgeries using sensor pattern noise. In: Security, steganography, and
  watermarking of multimedia contents VIII. SPIE (2006)

\bibitem{mensink2014costa}
Mensink, T., Gavves, E., Snoek, C.G.: Costa: Co-occurrence statistics for
  zero-shot classification. In: CVPR (2014)

\bibitem{nguyen2019multi}
Nguyen, H.H., Fang, F., Yamagishi, J., Echizen, I.: Multi-task learning for
  detecting and segmenting manipulated facial images and videos. In: IEEE BTAS
  (2019)

\bibitem{nguyen2019capsule}
Nguyen, H.H., Yamagishi, J., Echizen, I.: Capsule-forensics: Using capsule
  networks to detect forged images and videos. In: IEEE ICASSP (2019)

\bibitem{oquab2023dinov2}
Oquab, M., Darcet, T., Moutakanni, T., Vo, H., Szafraniec, M., Khalidov, V.,
  Fernandez, P., Haziza, D., Massa, F., El-Nouby, A., et~al.: Dinov2: Learning
  robust visual features without supervision. arXiv preprint arXiv:2304.07193
  (2023)

\bibitem{qian2020thinking}
Qian, Y., Yin, G., Sheng, L., Chen, Z., Shao, J.: Thinking in frequency: Face
  forgery detection by mining frequency-aware clues. In: ECCV (2020)

\bibitem{qiu2023learnable}
Qiu, Z., Hu, Y., Li, H., Liu, J.: Learnable ophthalmology sam. arXiv preprint
  arXiv:2304.13425  (2023)

\bibitem{radford2021learning}
Radford, A., Kim, J.W., Hallacy, C., Ramesh, A., Goh, G., Agarwal, S., Sastry,
  G., Askell, A., Mishkin, P., Clark, J., et~al.: Learning transferable visual
  models from natural language supervision. In: ICML (2021)

\bibitem{rossler2018faceforensics}
R{\"o}ssler, A., Cozzolino, D., Verdoliva, L., Riess, C., Thies, J.,
  Nie{\ss}ner, M.: Faceforensics: A large-scale video dataset for forgery
  detection in human faces. arXiv preprint arXiv:1803.09179  (2018)

\bibitem{swaminathan2008digital}
Swaminathan, A., Wu, M., Liu, K.R.: Digital image forensics via intrinsic
  fingerprints. IEEE TIFS  (2008)

\bibitem{tan2019efficientnet}
Tan, M., Le, Q.: Efficientnet: Rethinking model scaling for convolutional
  neural networks. In: ICML (2019)

\bibitem{wu2023medical}
Wu, J., Fu, R., Fang, H., Liu, Y., Wang, Z., Xu, Y., Jin, Y., Arbel, T.:
  Medical sam adapter: Adapting segment anything model for medical image
  segmentation. arXiv preprint arXiv:2304.12620  (2023)

\bibitem{zhang2023comprehensive}
Zhang, C., Liu, L., Cui, Y., Huang, G., Lin, W., Yang, Y., Hu, Y.: A
  comprehensive survey on segment anything model for vision and beyond. arXiv
  preprint arXiv:2305.08196  (2023)

\bibitem{zhou2014low}
Zhou, X., Yang, C., Zhao, H., Yu, W.: Low-rank modeling and its applications in
  image analysis. ACM Computing Surveys (CSUR)  (2014)

\end{thebibliography}

%




\end{document}